\documentclass[doc]{article}
\usepackage[utf8x]{inputenc}

\usepackage{subfigure} 
\usepackage{apacite}
\usepackage{amsmath}
\usepackage{multirow}
\usepackage{graphicx}

\title{Information content versus word length in natural language: A reply to Ferrer-i-Cancho and Moscoso del Prado Martin (2011)}
\author{Steven T. Piantadosi, Harry Tily, Edward Gibson}{

\begin{document}
\maketitle

\begin{abstract}
Recently, \citeA{ferrer2011information} argued that an observed linear relationship between word length and average surprisal \cite{piantadosi2011word} is not evidence for communicative efficiency in human language. We discuss several  shortcomings of their approach and critique: their model critically rests on inaccurate assumptions, is incapable of explaining key surprisal patterns in language, and is incompatible with recent behavioral results. More generally, we argue that statistical models must not critically rely on assumptions that are incompatible with the real system under study.
\end{abstract}

\section{Introduction}

One of the most famous properties of language, first studied by \citeA{zipf1936psychobiology}, is that frequent words are typically also short. Zipf offered a communicative theory for this property, under which lexicons have evolved to be efficient: words which must be re-used repeatedly should be short to minimize the effort of language users. Recently, we \cite[henceforth, PT\&G]{piantadosi2011word} demonstrated an improvement on Zipf's ideas. Under a more elaborate notion of communicative efficiency, word length should depend not on frequency, but on the typical \emph{amount} of information conveyed by a word. Efficient languages will convey information close to the channel capacity \cite{shannon1948mathematical} of human perceptual and cognitive systems. In this case, one should observe a linear relationship between a word's negative log probability (surprisal) and its length, in an attempt to keep the number of bits communicated per unit time 
roughly constant. Such a prediction is a lexical version of a now popular idea that choices made in language production also attempt to maintain a roughly uniform rate of information transmission \cite{genzel2002entropy,genzel2003variation,AylettTurk2004,levy2007speakers,jaeger2010redundancy}. PT\&G demonstrated across $10/11$ languages for which corpora were readily available that information content does predict word length better than frequency, both in total correlations and partial correlations. 

Recently, \citeA[henceforth F\&M]{ferrer2011information} argued that the roughly linear relation observed in PT\&G was not necessarily evidence of communicative efficiency. They prove that a model in which language is generated by choosing characters independently also shows a linear relationship between average information content and word length\footnote{Because in their model predictability reduces to frequency, their work replicates \citeA{miller1957some} (e.g. Equation 2) and extends it to the case of unequal letter probabilities.}. In such a system, words are generated by randomly typing characters, and occasionally hitting the ``space'' character to create a word boundary. It is intuitively not surprising that such a system would show the required linear relationship, since the probability of achieving a word of a given length $l$ will decrease geometrically in $l$, so the log probability scales linearly in $l$. F\&M furnish a mathematical proof for a more general version where all words must be 
longer than some minimum length $l_0$ and individual letters occur with arbitrary probabilities. Because random typing does not consider communicative efficiency, they argue that it is possible to achieve a linear relationship without any notion of communicative optimization\footnote{F\&M do not specify what they mean by communicative efficiency. For PT\&G and other UID work before them, language would be efficient if it tended to communicative bits of information at the channel capacity of human cognitive systems. Under this definition, F\&M's random typing model actually could be efficient.}. Though they do not say so explicitly, their paper implies that such simple statistical models should be treated as baselines, where any properties of language that hold in them are not expected to be the result of any interesting causal processes. F\&M's random typing model is one exemplar of a long history of \emph{monkey models} in psycholinguistics, so-called because they capture the Borel's process 
of ``a million monkeys typing on a million typewriters.'' Such models were first articulated in the study of language by \citeA{mandelbrot1953informational} and \citeA{miller1957some} in an attempt to account of the power law distribution of word frequencies \cite{zipf1936psychobiology}.

F\&M's model is essentially a toy model of language and does indeed exhibit a linear relationship between word length and information content. However, we argue that their model is, in some sense, too simplified for the strong claims they make. Here, we discuss some of the limitations of F\&M's model. After clarifying the main finding of PT\&G, we argue that the model's assumptions make it inherently unlike real human language, and therefore a poor choice for statistical comparison. We then review recent behavioral work indicating that PT\&G's results are not statistical artifacts. 

\subsection*{Random typing cannot explain the primary finding of PT\&G}

First---and perhaps most importantly---F\&M do not address the primary data point reported by PT\&G. Our main finding was \emph{not} a linear relationship. Instead, we focused on reporting that a word's average in-context surprisal predicted word length \emph{better than frequency predicts word length}. This pattern held in general for several different corpora, ways of measuring word length, and ways of estimating surprisal, and in partial correlations (e.g. surprisal partialing out frequency vs. frequency partialing out surprisal). Thus, PT\&G's measure of the average amount of information conveyed by a word was a more important determinant of word length than frequency---so much so that the partial effect of frequency was near zero in some languages. Indeed, our primary correlations reported were not even linear correlations, but nonparametric (although both give very similar results). 

In the random typing setup used by F\&M, frequency and our information measure are mathematically identical---a fact used in their derivations---so random typing could never find that information content was a better predictor of length than frequency. This both illustrates a limitation of F\&M's model, and provides evidence that it is a poor description of the statistical patterns in natural language. 

%

\subsection*{Independent data argues against random typing}

Beyond the fact that F\&M's analysis cannot address the reported differences between frequency and information content, it is worth considering its limitations when viewed as a statistical model of language. The model's generative process assumes that words are created by just happening to randomly sample their component pieces. This probabilistic scheme is what assigns words of varying lengths their varying probabilities.

But language generation does not work that way\footnote{\citeA{howes1968zipf} presented a similar critique to Miller's random typing model for deriving the Zipfian distribution of word frequencies: ``If Zipf's law indeed referred to the writings of `random monkeys,' Miller's argument would be unassailable, for the assumption he bases it upon are appropriate to the behavior of those conjectural creatures. But to justify his conclusion that people also obey Zipf's law for the same reason, Miller must perforce establish that the same assumptions are also appropriate to human language. In fact, as we shall see, they are directly contradicted by well-known and obvious properties of languages.''}. Instead, speakers \emph{know whole words}. This fact is not hard to establish psychologically or statistically. For instance, psychologically, speakers know word meanings and produce them in the correct context---they don't just happen to say words by randomly saying syllables or phonemes. Statistically, idealized models 
over sequences of characters infer not only the presence of words but the correct words themselves \cite{goldwater2006nonparametric,pearl2011online}. Though such models were proposed as language acquisition models, they equally serve as idealized data analysis models, demonstrating that the evidence provided by statistical dependencies between characters in natural language strongly favors the existence of words. 


The existence of words as psychological units undermines F\&M's primary point, that ``a linear correlation between information content and word length may simply arise internally, from the units making a word (e.g., letters) and not necessarily from the interplay between words and their context ... .'' If humans generate language by remembering entire words, rather than by sampling their component parts, then there is no necessary relationship between word length, frequency, and information content. In the real psychological system wordforms are memorized sequences, making their probability of generation is not longer intrinsically tied to their length. Indeed, it is hard to see what the behavior of models that critically rely on randomly generating word \emph{components} could tell us about a psychological system that does not work that way.

\subsection*{Random typing is not even a good statistical model}

It is still worth considering that even though the generative assumptions of random typing models are limiting, they still may provide a useful \emph{statistical} description of language. Unfortunately, a considerable amount of evidence has amassed that such models are poor statistical theories. \citeA[pg106-107]{baayen2001word} analyzes the predictions of a random typing model with respect to coarse properties of lexical systems, including word frequency distributions, frequency/length relationships, and neighborhood density. He finds that while a random typing model provides qualitative trends in the right directions, its quantitative fit is not very good and is eclipsed by the fit of other models such as a Yule-Simon model. Indeed, a considerable amount of work---confusingly, much of it by the first author of F\&M---has detailed the ways in which the output of random typing models are \emph{unlike} those found in natural languages, especially with respect to Zipf's law \cite{tripp1982zipfs,baayen2001word,
montemurro2001beyond,ferrer2002zipf,
ferrer2010random}. Other statistical properties of random texts have been found to be divergent from real language. For instance, \citeA{bernhardsson2011paradoxical} 
shows that random texts, but not natural texts, follow the statistics of Heaps' law, a growth pattern relating types and tokens to text sample size \cite{heaps1978information}. \citeA{cohen1997numerical} compares entropy-based measures on natural and random typing model texts, with the goal of finding metrics that best distinguish these texts. Our other work has detailed ways in which lexical systems are not only non-random, but specifically structured for communicative efficiency, in terms of ambiguity \cite{piantadosi2010communicative} or lexical properties such as stress \cite{piantado2009communicative}. Additionally, not all short, phonotactically possible words are used in language \cite{cohen2006why}, contrary to the predictions of a random typing model, but consistent with communicative theories based on entropy rate (PT\&G) or possibly the avoidance of confusable code words. In short, it is clear that random typing models don't even produce the correct detailed \emph{statistical} properties of 
language, although they may 
appear qualitatively similar with respect to some coarse-grained properties. 

\subsection*{The importance of a model's key assumptions}

The implication of F\&M (and before them, \citeA{miller1963finitary}) is that even though random typing models are implausible descriptions of the generative process of language and poor statistical theories, they still provide a null hypothesis which should be considered in the course of scientific theorizing. Properties of language that are also exhibited by random typing models should be looked on cautiously, as phenomena which likely have a trivial and uninteresting cause. In contrast, we believe that it is a fallacy to think that the fact a random typing model exhibits a linguistic property should cast any doubt on alternative theories, such as those proposed by PT\&G and Zipf. The hypothesis of random typing---and all models like them---have already been disproven by other sources of evidence like the statistical and psychological existence of words as memorized units. 

We find this point interesting because it raises a difficult issue for modeling. All models are inaccurate in that they do not exactly mirror the ``real'' process happening in the world. Most models do or should attempt to make their key assumption analogous to the key causal process at play in the world. Thus, changing ``good'' models to make them more realistic will not break their core predictions and properties. But F\&M's model is different: it cannot be given knowledge of words---like people have---without destroying the behavior F\&M aim to explain. The key assumption of generating words by happening to choose their components make the model critically \emph{unlike} people in terms of representation, processing, and knowledge of language. 

\subsection*{Independent evidence for PT\&G's optimization process}

Aside from discussions of the plausibility of various models, there are good independent reasons for rejecting F\&M's assertion that the relationship observed by PT\&G is a statistical artifact. PT\&G posited that the observed relationships might result from lexicalization of phonetic reduction \cite<e.g.>{lieberman1963some,jurafsky2001evidence}. It is well-known that speakers shorten or reduce syllables in predictable locations and PT\&G's findings plausibly result from these speech production factors being integrated into lexical representations. If a word is used in predictable locations, it will be reduced, and eventually might be learned to be its shorter form, giving a relationship between word length and predictability. 

Second, there are independent behavioral studies showing that speakers actually \emph{do} prefer short forms of words in predictable contexts, exactly as PT\&G's theory would predict. \citeA{mahowald2012info} gave people a choice between two synonymous pairs (e.g. ``chimp''/``chimpanzee'') in either predictive or non-predictive contexts. They found that people preferred the the short form when the word was predictable and the long form was it was not. These kind of behavioral tendencies have also been examined in corpus research by \citeA{frank2008speaking}, who showed that contracted forms (``do not'' / ``don't'') occur more frequently in predictive contexts. Such behavior is predicted by PT\&G, but not explainable with F\&M's view that the relationship between information content and word length is a statistical artifact.

\subsection*{Conclusion}

Random typing models provide an interesting case study for considering what modelers should want from models. Good models do not simply exhibit the correct surface statistics; good models capture the right core assumptions of the system under study, and show how the observed properties of the system result from those properties. It is not informative to show that other assumptions could also lead to the observed behavior, \emph{if} those other assumptions are demonstrably not at play.

This means that one is not free to study \emph{any} conceivable statistical process and conclude that it is relevant for understanding how language works. Models under consideration must respect what is independently known about the system under study. Since words are actively \emph{chosen} by language users to convey a meaning, there is no point to studying models for which the uttered word is generated according to some statistical properties of the wordform itself---that is the wrong causal direction. As such, results about random typing models only apply to systems that are critically unlike human language in terms of the structure of language, knowledge of words, and the transmission of meaningful information.

\bibliographystyle{apacite}
\bibliography{AllCitations}

\end{document}